 \documentclass{colt} 

\title[Quantum Bandits]{Quantum Bandits}
\usepackage{times}

 \coltauthor{\Name{Balthazar Casal\'e} \Email{balthazar.casale@lis-lab.fr}\\
 \addr Aix-Marseille Universit{\'e}, CNRS, LIS, Marseille, France\AND
  \Name{Giuseppe Di Molfetta} \Email{giuseppe.dimolfetta@lis-lab.fr}\\
  \addr Aix-Marseille Universit{\'e}, CNRS, LIS, Marseille, France and Quantum Computing Center, Keio University, 3-14-1 Hiyoshi, Kohoku, Yokohama, 223-8522 Japan\AND
  \Name{Hachem Kadri} \Email{hachem.kadri@lis-lab.fr}\\
  \addr Aix-Marseille Universit{\'e}, CNRS, LIS, Marseille, France
\AND
\Name{Liva Ralaivola} \Email{l.ralaivola@criteo.com}\\
\addr Criteo AI Lab, Criteo, Paris%
}

\usepackage[utf8]{inputenc} 
\usepackage[T1]{fontenc}    
\usepackage{hyperref}       
\usepackage{url}            
\usepackage{booktabs}       
\usepackage{amsfonts}       
\usepackage{nicefrac}       
\usepackage{microtype}      
\usepackage{lipsum}
\usepackage{ulem}           

\usepackage[ruled]{algorithm2e}
\usepackage{amssymb}
\usepackage{physics}
\usepackage{graphicx}

\newcommand{\Hi}{\mathcal{H}}

\newcommand{\Id}{\mathbb{I}}

\DeclareMathOperator*{\argmax}{\arg\,max}

\definecolor{MyDarkBlue}{rgb}{0.05,0.,0.8} 
\newcommand{\addHK}[1]{{\color{black}  #1}}

\begin{document}

\maketitle

\begin{abstract}

We consider the quantum version of the bandit problem known as {\em best arm identification} (BAI). 
We first propose a quantum modeling of the BAI problem, which assumes that both the learning agent and
the environment are quantum; we then propose an algorithm based on quantum amplitude amplification
to solve BAI. We formally analyze the behavior of the algorithm  
on all instances of the problem and we show, in particular, that it is able to get the optimal solution quadratically faster than what is known to hold in the classical case. 
\end{abstract}

\medskip

\section{Introduction}
\label{sec:intro}

Many decision-making problems involve learning by interacting with the environment and observing what rewards result from these interactions.
In the field of machine learning, this line of research falls into what is referred as 
reinforcement learning (RL), and algorithms to train artificial agents that interact
with an environment have been studied extensively~\citep{Sutton1998,KLMSurvey,BertsekasTsitsiklis96}. 
We are here interested in the best arm identification (BAI) problem
from the family of bandit problems, which pertains the set of RL problems where the interactions with the environment give 
rise to immediate rewards and where long-term planning is unnecessary (see the survey of \citealp{lattimore_szepesvari_2020}). More precisely,
we are interested in a quantum version of the BAI problem, for which
we design a quantum algorithm capable to solve it.

Quantum machine learning is a research field at the interface of quantum computing and machine learning where the goal is to use quantum
computing paradigms and technologies to improve the speed and performance of learning algorithms~\citep{wittek2014quantum, biamonte2017quantum, ciliberto2018quantum, schuld2018supervised}.
A fundamental concept in quantum computing is {\em quantum superposition}, which is
the means by which quantum algorithms like that of \cite{grover1996fast} ---one
of the most popular quantum algorithm---  succeeds
in solving the problem of finding one item from an unstructured database of $N$ items in time $O(\sqrt{N})$, so beating the classical  $O(N)$ time requirement.
Recent works have investigated the use of Grover's quantum search algorithm to enhance machine learning and have proved 
its ability of providing non-trivial improvements not only in the computational complexity but also in the statistical performance of these models~\citep{aimeur2013quantum, wittek2014quantum,kapoor2016quantum}. Beyond Grover's algorithm, quantum algorithms for linear algebra, such as quantum matrix inversion
and quantum singular value decomposition, were recently proposed and used in the context of machine learning~\citep{rebentrost2014quantum, kerenidis2017quantum}. 
%
Works on quantum reinforcement learning are emerging \citep{dong2008quantum, naruse2015single, dunjko2016quantum, lamata2017basic}, and our paper aims at providing a new piece
of knowledge in that area, by bringing two contributions: i) a formalization of the best arm identification problem in a quantum setting, and ii) a quantum algorithm to solve this problem that is quadratically faster than classical ones.

\addHK{Quantum machine learning research can be classified into four categories depending on whether the data, the learner, both, or none are quantum~\citep{aimeur2006machine,dunjko2018machine}. Our work deals with the BAI problem when both the agent and the environment are quantum systems, and so falls into the Quantum-Quantum~(QQ) setting.
Although less studied, the QQ approach is particularly attractive because it  would allow the exploitation of the full potential of quantum technologies in machine learning.
In this setting, the interaction can be fully quantum, and the agent and the environment may become entangled~\citep{dunjko2016quantum}.
Recent progress in reinforcement learning has achieved very impressive results in games~\citep{mnih2015human} and robotics~\citep{levine2016end}.
The training process of these models is often done in a computer simulated environment, as it would require too much agent-environment interactions to be done with a physical system in a reasonable amount of time. 
Performing such simulations on a quantum computer or simulator should give rise to environment's internal states that are naturally quantum.
The internal state of the environment may be hidden from the agent, and considering quantum interactions between the agent and the environment would lead to more efficient learning. This motivates the setting we
are interested in: quantum agents and quantum environments.}

The paper is organized as follows. In Section~\ref{sec:bandit}, we formulate the best arm identification (BAI) problem, briefly review the upper confidence bound, and illustrate how it can be used to solve the BAI problem. In Section~\ref{sec:quantum}, we describe the quantum amplitude amplification, at the core of Grover's algorithm,  which forms the basis of our approach. Our main results are in Section~\ref{sec:qubandit}: we provide our quantum modeling of the BAI problem, which assumes that both the learning agent and the environment are quantum; and then we proposes an algorithm based on quantum amplitude amplification to solve BAI, that it is able to get the optimal solution quadratically faster than what is known to hold in the classical case. Section~\ref{sec:conclusion} concludes the paper.

\section{Best Arm Identification}
\label{sec:bandit}
\subsection{Stochastic Multi-Armed Bandits and the BAI Problem}

Bandit problems are RL problems where it is assumed an agent evolves in an
environment with which it can interact by choosing at each time step an action (or arm),
each action taken providing the agent with a reward, which values the 
quality of the chosen action (see function $f$ below, and more generally, \citealp{lattimore_szepesvari_2020}). 

The bandit problem we want to study from a quantum point of view 
is that of best arm identification from stochastic multi-armed bandits
\citep{best_arm_ident}.  It comes with the following assumptions: 
the set $X$ of actions is finite and discrete, with $|X|=N$, and when 
action $x_t$ is chosen at time $t$ then the reward $r_t$ depends
upon the independent realisation (called $y_t$ afterwards) of a random variable
distributed according to some unknown (but fixed) law $\nu_{x_t}.$ The
 BAI problem is to devise a strategy of action selection for the agent such that,
 after a predefined number $T$ of interactions, the agent is able to
 identify the best action with the best possible guarantees.

We may go one step further in the formal statement of the problem and, 
in the way, use a modelling that is both in line with the classical BAI 
problem and suitable for its quantum extension. In particular, in order to take
the unknown distributions $\nu_x,\,x\in X$, we will explicitly introduce
$Y$, the set of all possible internal states $y_t$ of the environment ---this
notion of internal state of the environment is uncommon in the classical bandit
literature.
The agent's action $x_t$ sets the internal state of the environment to
$y_t$, which is a random draw from distribution $\nu_{x_t}$, unknown to the agent. 
The agent then receives a reward $r_t = f(x_t,y_t)$, indicating the fit of action $x_t$ with the state of the environment; we
here assume that $f$ can only take values in $\{0,1\}$ ---this corresponds to the classical case where
the reward $r_t$ is drawn according to a Bernoulli distribution of unknown parameter
$\theta_{x_t}\in[0,1]$.
With these assumptions, the average reward associated with
action $x$ is 
\begin{equation}
    a_x = \sum_{y \in Y} \nu_x(y) f(x,y) ,
    \label{eq:average_reward_x}
\end{equation}
and we may define the optimal action $x^*$ as
\begin{equation}
    x^* = \argmax_{x \in X} a_x,
    \label{eq:optimal_action}
\end{equation}
 and $a^* = a_{x^*}$ the mean reward of the optimal action. 
 After $T$ interactions with the environment, the agent will choose an action $\tilde x_T$ as its recommendation (see Algorithm~\ref{alg:bai}). The quality of the agent's decision $\tilde x_T$ is then evaluated as the {\em regret} $a^*-a_{\tilde{x}_T}$, i.e. the difference between $a^*$ the mean reward of optimal action $a^*$ and $a_{\tilde x_T}$ the mean reward of the recommended action.

\begin{algorithm}[t]
	\KwData{A number of rounds $T$}
	\KwResult{$\tilde x_T$ a recommended action}
	
	\For{$t\gets1$ \KwTo $T$}{
		{the agent chooses the action $x_t$ \;\newline
		the environment picks an internal state $y_t$ following $\nu_{x_t}$ \;\newline
		the agent perceives the reward $r_t = f(x_t,y_t)$ \;}
	}
	the agent return $\tilde x_T$ the recommended action\;
	\caption{The best arm identification problem \label{alg:bai}}
\end{algorithm}

Let us elaborate further on the regret; let 
\begin{equation}
    \Delta_x = a^* - a_x
    \label{eq:delta}
\end{equation}
be the difference between the value of the optimal action and the value of action $x$. If the agent recommends the action $x$ with probability $P_T(x)$ after $T$ rounds, then the average difference between the value of its recommendation and the value of the optimal action is 
\begin{equation}
    R_T = \sum_{x \in X} P_T(x) \Delta_x,
    \label{eq:average_regret}
\end{equation} 
which is the {\em average regret} after $T$ iterations of the agent's 
strategy. Our goal is to find an action selection strategy for which the value of $R_T$ decreases quickly as the value of $T$ increases.

If $e_T = 1 - P_T(x^*)$ is the probability that the agent does not recommend the best action after $T$ iterations, then, as $\forall x \in X, \Delta_x \leq 1$, the (average) regret is so that $R_T < e_T$. In the following, we recall how a tight upper bound for $e_T$ can 
be derived.

\subsection{Upper Confidence Bound Exploration-based strategy}

Part of the difficulty in the BAI problem comes from the fact 
that the value of each action is the mean of random variable
that depends on an unknown probability distribution. 
The only way for an agent to estimate the value $a_x$ of action $x$ 
is to repeatedly interact with the environment to obtain a sample of
rewards associated to $x$. Thus, a good strategy needs to find a
balance between sampling the most promising actions, and sampling the
actions for which we lack information. 
The Upper Confidence Bound Exploration (UCB-E) depicted in Algorithm~\ref{alg:ucbe}, first described in \cite{best_arm_ident}, is an efficient strategy to solve the best arm identification problem. It is based on a very well known and used family of UCB strategies \citep{band_effic, band_fini}, which were proven to be optimal for solving the multi-armed bandit problem \citep{band_thompson}. 

Let $\Omega_x(T)$ be the set of rounds for which the agent picked action $x$ until time $T$, and 
\begin{equation}
    \tilde{a}_x(T) = \frac{1}{\abs{\Omega_x(T)}}\sum_{t \in \Omega_x(T)} r_t
    \label{eq:empirical_reward_x}
\end{equation}
be the empirical average of the reward for action $x$.
We know from \cite{oth_hoeffding} that $a_x$ and $\tilde{a}_x(T)$ are tied by the relation $$\mathbb{P}(\abs{\tilde{a}_x(T) - a_x} > \epsilon) < 2\exp(-2\epsilon^2\abs{\Omega_x(T)}).$$ 
This means that, for all $\delta \in [0,1]$, there is a range of value centered around $\tilde{a}_x(T)$ in which $a_x$ lies with probability at least $1-\delta$. The more the agent interacts with the environment with action $x$, the smaller this range of values is. The principle behind UCB is to choose, at each iteration, the action $x$ for which the upper bound of this range is the highest. 

\begin{algorithm}[t]
	\KwData{a number of trials $T$\; \newline an exploration parameter $p$}
	\KwResult{$\tilde x_n$ a recommended action}
	
	let $B_{x,t} = \tilde{a}_x(t) + \sqrt{\frac{p}{t-1}}$\;\newline
	\For{$t\gets1$ \KwTo $T$}{
		{the agent chooses the action $x_t \in \argmax_{x \in X} B_{x,t}$\;\newline
		the environment picks an internal state $y_t$ according to $\nu_{x_t}$\;\newline
		the agent perceives the reward $r_t = f(x_t,y_t)$ \;\newline
		the agent updates the values of $B_{x,t}$ to take $r_t$ into account \;}
	}
	the agent return $\tilde x_T = \argmax_{x \in X} \tilde{a}_x(T)$ \;
	\caption{UCB-E algorithm \label{alg:ucbe}}
\end{algorithm}

\cite{best_arm_ident} showed that UCB-E admits the following upper bound on $e_T$, when the exploration parameter $p$ is well tuned :
$$
e_T <  2T N\exp\left(- \frac{T - N}{18H_1}\right),\;\text{where } H_1=\sum_{x\in X\backslash\{x^*\}} \frac{1}{\Delta_x^2}.
$$
 From this inequality, we can deduce a lower bound of the number of iterations to recommend the optimal arm with probability at least $1 - \delta$, for any $\delta\in(0,1)$:
$$
e_T < \delta \Rightarrow T >  18H_1\ln\left(\frac{2N}{\delta}\right) + N.
$$
The quantum modelling and accompanying algorithm proposed 
in this paper come with a theoretical result that quadratically
improves this bounds.

\section{Quantum Amplitude Amplification}
\label{sec:quantum}
If we dispose of an unstructured, discrete set $X$ of $N$ elements and we are interested in finding one marked element $x_0$, a simple probability argument shows that it takes an average of N/2 (exhaustive) queries to find the marked element. While it is well known that $O(N)$ is optimal with classical means, \cite{grover1996fast} proved that a simple quantum search algorithm speeds up any brute force $O(N)$ problem into a $O(\sqrt{N})$ problem. This algorithm comes in many variants and has been rephrased in many ways, including in terms of resonance effects~\citep{grover1996fast} and quantum walks~\citep{childs2004spatial,guillet2019grover}. The principle behind the original Grover search algorithm is the amplitude amplification \citep{ampli_esti,grover1998quantum} in contrast with the techniques called probability amplification used in classical randomized algorithms. 

In the classical case it is known that, if we know the procedure which verifies the output, then we can amplify the success probability $n$ times, and the probability to recover the good result is approximately $np$ where $p$ is the probability to return the searched value. Thus in order to amplify the probability to $1$ we need to multiply the runtime by a factor $1/p$. In the quantum case, the basic principle is the same and we amplify amplitudes instead of probabilities. Grover's algorithms and all its generalisations have shown that in order to achieve a maximum probability close to 1, we amplify for a number of rounds which is  $O(\sqrt{1/p})$, then quadratically faster {than} the classical case. Before we show how to apply this result to the best arm identification problem, let us briefly recall how the amplitude-amplification algorithms works. 
First, we need to introduce a $N$-dimensional state space $H$, which can be supplied by $n=\log_2 N$ qubits, spanned by the orthonormal set of states $\ket{x}$, with $x\in X$. In general, we say that, after the application of an arbitrary quantum operator, the probability to find the marked element $x_0$ is $p$, where this element is a point in the domain of a generic Boolean function $f:\{0,1\}^n \rightarrow \{0,1\}$ such that $f(x_0) =1$. This function induces a partition of {$\Hi$} into two subspaces, {$\Hi_1$ and $\Hi_0$}, and each of them can be seen respectively as the good subspace spanned by the set of basis states for which $f(x)= 1$ and the bad subspace, which is its orthogonal. Any arbitrary state $\ket{\Psi}$ belonging to {$\Hi$} can be decomposed on the basis $\{\ket{\Psi_1}, \ket{\Psi_0}\}$ as follows
\[
\ket{\Psi} = \sin\theta \ket{\Psi_1} + \cos\theta\ket{\Psi_0},
\]
where $\{\ket{\Psi_1}, \ket{\Psi_0}\}$ are the normalised projections of $\ket{\Psi}$ in the two subspaces {$\Hi_1$ and $\Hi_0$}:
\[
\ket{\Psi_1} = \frac{1}{\sqrt{p}}\sum_{f(x)=1}\alpha_x \ket{x}, \quad 
\ket{\Psi_0} = \frac{1}{\sqrt{1-p}}\sum_{f(x)=0}\alpha_x \ket{x},
\]
{ where $\alpha_x$ is a complex number} and $\sin\theta = \sqrt{p}$ denotes the probability that measuring $\ket{\Psi}$ produces a marked state (for which $f(x)= 1$). 
In general terms, one step of the algorithm is composed by two operators: (i) the oracle, as in the original Grover results; (ii) and the generalised Grover diffusion operator. The oracle $O_f$ is built using $f$ and reads:
\[
{O_f\ket{x} = (-1)^{f(x)}\ket{x},}
\]
which essentially \textit{marks} the searched state with minus sign. The diffusion operator is defined as:
\[
R_\Psi = AS_0A^{-1} = 2\ket{\Psi}\bra{\Psi} - { \Id},
\]
where $S_0 = 2\ket{0}\bra{0} - {\Id}$ is the usual reflection operator around $\ket{0}$ and $\ket{\Psi}= A\ket{0}$. 
The composition of both operators leads to one evolution step of the amplitude-amplification algorithm:
\[
Q = R_\Psi O_f.
\]
Notice that when $A = H^{\otimes n}$, the Walsh-Hadamard transform, the above algorithm reduces to the original Grover algorithm, where the initial state is an uniform superposition of states.
The repetitive application of $Q$ after $n$ iterations leads to:
\begin{equation}
\label{eq:qaa}
Q^n \ket{\Psi} = \sin((2n+1)\theta)\ket{\Psi_1} + \cos((2n+1)\theta)\ket{\Psi_0}.
\end{equation}
As in the Grover algorithm for $n \approx \frac{\pi}{4\theta}$ and $\theta \ll 1$, { the number of call to $Q$ needed to find the desired element is in $O(\frac{1}{\sqrt{p}})$}, leading { to a }quadratic speedup over classical algorithms.

\section{Quantum Best Arm Identification}
\label{sec:qubandit}
Efficiently Solving the best arm identification problem is generally limited by the amount of information the agent needs to recover from a single interaction with the environment. This is also the case in the unstructured classical search problem, as a single call to the indication function $f$, the oracle, gives us information on a single element of the set. In general terms, the idea is to apply the same basic principle of the amplitude-amplification quantum algorithm to the best arm identification problem, where the reward function introduced in Section \ref{sec:bandit} now plays the role of the oracle. Indeed, in the same way that the boolean function $f$ in a searching problem \textit{recognises} whether $x$ is the marked element we are looking for, the reward $r_t = f(x_t,y_t)$, indicates  whether $\{x_t,y_t\}$ corresponds to a desirable outcome (in that case, $f(x_t,y_t)=1$) or not (then $f(x_t,y_t)=0$), where $x_t$ is the action of the agent and $y_t$ the state of the environment. Thus, our strategy in the following is to apply the amplitude-amplification quantum algorithm to recover the desirable outcome, i.e., the optimal action of the agent. 

In order to properly apply the above quantum strategy, we define a composite Hilbert space $\Hi = \Hi_X\otimes\Hi_Y$, where $\Hi_X$ is the space of the quantum actions of the agent, spanned by the orthonormal basis $\{\ket{x}\}_{x \in X}$ and $\Hi_Y$ is the space of the quantum environment states, spanned by the orthonormal basis $\{\ket{y}\}_{y \in Y}$. All vector $\ket{\Psi}$, representing the whole composite system, decomposes on the basis $\{\ket{x y}\}_{x \in X, y \in Y}$. Notice that in the classical context, the agent's action sets the internal state of the environment to $y_t$, according to a random distribution $\nu_{x_t}$, which is unknown to the agent. A straightforward way to recover the same condition, is to prepare the state of the environment in a superposition {$\ket{\psi_x} = \sum_{y \in Y} \sqrt{\nu_x(y)} \ket{y}$}, where $\nu_x(y)$ depends on the action $x$ chosen by the agent. This is achieved preparing the initial state of the environment as follows:
\[
\forall x \in X,\hspace{0.4cm} O_e\ket{x 0} = {\ket{x \psi_x} : \abs{\braket{y}{\psi_x}}^2 = \nu_x(y) },
\]
where $O_e$ is a unitary operator acting on the composite Hilbert space $\Hi$. 
Moreover, the initial state of the agent is prepared in an arbitrary superposition state, applying an unitary operator $A$ on the state space of the agent $\Hi_X$:
\[
A\ket{0}={\ket{\phi} = \sum_{x \in X} { \alpha_x } \ket{x}}.
\]
Once the initial state is prepared, we build the oracle $O_f$ on the composite Hilbert space of the agent and the environment, the action of which is:
\[
\forall x\in X, y\in Y, \hspace{0.4cm}O_f\ket{xy} = 
\begin{cases}
-\ket{x y} \text{ if $f(x,y) = 1$,} \\
\ket{x y} \text{ otherwise}.
\end{cases}
\]
As for a search problem, we propose a quantum procedure that allows us to a find the optimal action (for which $r_t=f(x_t,y_t)=1$) using $O(1/\sqrt{p})$ application of $O_f$, with probability approaching 1. The quantum amplitude amplification algorithm and its analysis is then reminiscent of what was presented in Section \ref{sec:quantum}. One round of the algorithm is defined by the composition of the above three operators and the resulting algorithm QBAI (Quantum Best Arm
Identification) is depicted in Algorithm~\ref{alg:qbai}.
\addHK{As shown in Algorithm~\ref{alg:qbai}, our strategy is based on applying, at each iteration, the operator $G$, computed from $O_e$, $O_f$ and $A$. It is worth noting that although $G$ does not vary as a function of time/iteration, our strategy is able to take into account the reward at each time step. This is achieved by means of the environment's internal state which can be in a quantum superposition that evolves with time according to the reward obtained after performing an action.
}
\begin{algorithm}[t]
	\KwData{a unitary operator $A$ acting on $\Hi_X$\;\newline
	a unitary operator $O_e$ acting on the composite system agent-environment\;\newline
		$n$ number of rounds\;}
	\KwResult{the recommended action $\tilde x_n$}
	{prepare a quantum register to the state $\ket{00}$ \;\newline
	apply $O_e(A \otimes \Id_e)$ to the state of the register \;\newline
	}
	\For{$t\gets1$ \KwTo $n$}{
		apply $G = (O_e(A \otimes \Id)){ (S_0^{(X)} \otimes S_0^{(Y)})}(O_e(A \otimes \Id))^{-1}O_f$ to the state of the register \;
	}
	return $\tilde x_n$
	\caption{Quantum Best Arm Identification (QBAI) \label{alg:qbai}}
\end{algorithm}

{ Defining $\ket{\Psi} = O_e(A \otimes \Id) \ket{00}$,} iterating $n$ times the { \sout{above algorithm} operator $G$ we recover }


\[
G^n\ket{\Psi} = \sin((2n+1)\theta)\ket{\Psi_1} + \cos((2n+1)\theta)\ket{\Psi_0},
\]
which is of the same form of Equation~\ref{eq:qaa}, where now $\{\ket{\Psi_1}, \ket{\Psi_0}\}$ are the normalised projections of $\ket{\Psi}$ in the two subspaces {$\Hi_1$ and $\Hi_0$}, respectively the good subspace spanned by the set of basis states for which $r_t=f(x,y)= 1$ and the bad subspace, which is its orthogonal.
We know from Section~\ref{sec:quantum}, that to recover the optimal action we need to maximise the sinus. Let us choose an alternative, but equivalent, path. Let { us } compute the recommendation probability {$P_n(x) = \sum_{y \in Y} |\bra{xy}G^n\ket{\Psi}|^2$}. After a straightforward computation and few simplifications, it results:
	\[
	P_n(x) = \abs{\bra{x}A\ket{0}}^2 ( 1 + (a_x - p)C(p,n)),
	\]
where $C(p,n) = \frac{\sin((2n+1)\theta)^2 - p}{p(1 - p)}$, $p=\sin(\theta)^2$ and  $a_x =\sum_{y : f(x,y)=1} \abs{\braket{y}{\psi_x}}^2$. The recommendation probability $P_{n}(\tilde x)$ for the optimal action $\tilde x$ is then recovered when $\sin((2n+1)\theta)^2 = 1$, i.e. when $n \approx \frac{\pi}{4}\sqrt{1/p} - \frac{1}{2}$. 

Summarizing the results so far:

\begin{theorem}
\label{th:main_theorem}
	The probability $P_{n}(\tilde x)$ that QBAI will recommend the optimal action $\tilde x$ is maximized when $n \approx \frac{\pi}{4}\sqrt{1/p} - \frac{1}{2}$. It follows that {$P_{n}(\tilde x ) = \abs{\bra{\tilde x }A\ket{0}}^2\frac{a^*}{p}$}.
\end{theorem}

In order {to compare} this result with the classical bounds, we need to define $A$. For sake of simplicity, let consider $A$ so that $\forall x \in X, \abs{\bra{x}A\ket{0}}^2 = \frac{1}{N}$, which translates in $p = \mathbb{E}_X[a_x]$. From Theorem 1, we need
\[
n = \frac{\pi}{4}\sqrt{\mathbb{E}_X[a_x]^{-1}} - \frac{1}{2}
\]
rounds to recommend the optimal action with probability $1 - (1 - \frac{a^*}{N\mathbb{E}_X[a_x]})$.
Let us recall that UCB-E needs at least $ 18H_1\ln{(\frac{2N}{\delta})} + N$ rounds to recommend the optimal action with the same probability. The ratio between { \sout{both probabilities} both number of rounds} is of order $O(\sqrt{\mathbb{E}_X[a_x]}H_1\ln(\frac{2N^2\mathbb{E}_X[a_x]}{N\mathbb{E}_X[a_x] - a^*})+ \sqrt{\mathbb{E}_X[a_x]}N)$. In the case $\mathbb{E}_X[a_x] > \frac{1}{N}$, then $\sqrt{\mathbb{E}_X[a_x]}N > \sqrt{N}$ and the complexity gain for the quantum algorithm results quadratic in respect of the number of actions. Otherwise, since $H_1 > (N-1){a^*}^{-2}$, we get that $\sqrt{\mathbb{E}_X[a_x]}H_1 > {a^*}^{- \frac{3}{2}}\sqrt{N}$, and the speedup is once again quadratic in respect of the number of actions. This result is sufficient to prove that QBAI is quadratically faster than a classical algorithm to recommend the optimal arm with probability at least { $\frac{a^*}{N\mathbb{E}_X[a_x]} = (N - \sum_{x \neq x^*} \Delta_x/a^*)^{-1}$. }

{
We know from Theorem~\ref{th:main_theorem} that QBAI cannot identify the best arm with better probability without modifying the operator $A$ during the learning process. As such, QBAI does not allow one to identify the best action with arbitrary small margin of error, as can be done in the classical approach.
However, because it is able to attain the same level of confidence in fewer interactions with the environment than classical strategies, it is reasonable to think that an algorithm based on QBAI could identify the best action with arbitrarily small margin of error while keeping a quantum advantage. Devising such an algorithm is out of the scope of this paper and we leave this possibility for future research.
}
\section{Conclusion}
\label{sec:conclusion}

We studied the problem of Best Arm Identification (BAI) in a quantum setting. We proposed a quantum modeling of this problem when both the learning agent and the environment are quantum. We introduced a quantum bandit algorithm based on quantum amplitude amplification to solve the quantum BAI problem and showed that is able to get the optimal solution quadratically faster than what is known to hold in the classical case. Our results confirm that quantum algorithms  can have a significant impact on reinforcement learning and  open up new opportunities for more efficient bandit algorithms.

Our aim with this paper has been to provide a direct application of amplitude amplification to the best arm identification problem, and to show that it exhibits the same behavior it did in other problems of the same nature in term of efficiency.
It has been proposed a direct quantum analogue of the  multi-armed bandit problem, and an analytical proof that amplitude amplification can find the best action quadratically faster than the best known classical algorithm with respect to the number of actions. Future extensions of this work might include the following topics: (i) Could this algorithm be adapted to recommend the optimal action with arbitrarily small margin of error? (ii) Can it be possible to treat the case where the reward function have value in $\mathbb{N}$? (iii) Can this algorithm be adapted to solve more complex decision making problems? (iv) Can it be proven or disproven that amplitude amplification is optimal for this problem, as it is for other unstructured search problems?

\section*{Acknowledgements}
 This work has been funded by the
 French National Research Agency~(ANR) project QuantML~(grant number ANR-19-CE23-0011), the P{\'e}pini{\`e}re d'Excellence 2018, AMIDEX fondation, project DiTiQuS, and the ID \#60609 grant from the John Templeton Foundation, as part of the "The Quantum Information Structure of Spacetime (QISS)" Project.

\bibliography{ref_quantumbandits.bib}

%
%
%
%

\end{document}